\documentclass[10pt,twocolumn,letterpaper]{article}

\usepackage[pagenumbers]{cvpr}

\usepackage{graphicx}         
\usepackage{xcolor}           
\usepackage{caption}          

\usepackage[table]{xcolor}
\usepackage[dvipsnames]{xcolor}
\usepackage[most]{tcolorbox}
\usepackage{tabularx}
\usepackage{fancyvrb}
\usepackage{utfsym}
\usepackage{cuted}
\usepackage{capt-of} 

\definecolor{codegray}{gray}{0.95}

\definecolor{baselinecolor}{gray}{.9}

\definecolor{navyblue}{HTML}{0071BC}
\definecolor{hotpink}{HTML}{FF0080}

\definecolor{oai-white}{HTML}{FFFFFF}
\definecolor{oai-black}{HTML}{000000}
\definecolor{oai-red}{HTML}{FF4500}
\definecolor{oai-green}{HTML}{51DA4C}
\definecolor{oai-blue}{HTML}{0000FF}
\definecolor{oai-yellow}{HTML}{FFF639}
\definecolor{oai-magenta}{HTML}{FF45FF}
\definecolor{oai-cyan}{HTML}{00FFFF}
\definecolor{oai-orange}{HTML}{FE7600}
\definecolor{oai-violet}{HTML}{8A2BE2}
\definecolor{oai-brown}{HTML}{A0522D}
\definecolor{oai-green-050}{HTML}{F4FFF4}
\definecolor{oai-green-100}{HTML}{E9FFE8}
\definecolor{oai-green-200}{HTML}{D9FFD8}
\definecolor{oai-green-300}{HTML}{C9FFC7}
\definecolor{oai-green-400}{HTML}{A6FFA3}
\definecolor{oai-green-500}{HTML}{7CF178}
\definecolor{oai-green-600}{HTML}{51DA4C}
\definecolor{oai-green-700}{HTML}{3FA93B}
\definecolor{oai-green-800}{HTML}{2D712A}
\definecolor{oai-green-900}{HTML}{193718}
\definecolor{oai-gray-000}{HTML}{FFFFFF}
\definecolor{oai-gray-100}{HTML}{FAFAFA}
\definecolor{oai-gray-200}{HTML}{F5F5F5}
\definecolor{oai-gray-300}{HTML}{E5E5E5}
\definecolor{oai-gray-400}{HTML}{FFB7A4}
\definecolor{oai-gray-500}{HTML}{CDCDCD}
\definecolor{oai-gray-600}{HTML}{A8A8A8}
\definecolor{oai-gray-700}{HTML}{747474}
\definecolor{oai-gray-800}{HTML}{393939}
\definecolor{oai-gray-900}{HTML}{000000}
\definecolor{visual}{HTML}{A50E0E}       
\definecolor{linguistic}{HTML}{174EA6}   
\definecolor{relational}{HTML}{E37400}   
\definecolor{egocentric}{HTML}{0D652D}

\makeatletter
\renewcommand\paragraph{\@startsection{paragraph}{4}{\z@}%
  {0.5ex \@plus 0.2ex \@minus 0.1ex}
  {-1em}
  {\normalfont\normalsize\bfseries}}%
\makeatother

\definecolor{cvprblue}{rgb}{0.21,0.49,0.74}
\definecolor{natureblue}{RGB}{235, 246, 255}
\usepackage[pagebackref,breaklinks,colorlinks,allcolors=cvprblue]{hyperref}

\def\paperID{****}
\def\confName{CVPR}
\def\confYear{2026}

\title{EagleVision: A Dual-Stage Framework with BEV-grounding-based Chain-of-Thought for Spatial Intelligence}

\def\authorBlock{%
    Jiaxu Wan\textsuperscript{1,2,$\dagger$,$\clubsuit$}\,,
    Xu Wang\textsuperscript{1,$\dagger$,$\diamondsuit$}\,,
    Mengwei Xie\textsuperscript{1}\,,
    Hang Zhang\textsuperscript{1}\,,\\
    Mu Xu\textsuperscript{1}\,,
    Yang Han\textsuperscript{3}\,,
    Ding Yuan\textsuperscript{2,4}\,,
    Hong Zhang\textsuperscript{2,5}\,,
    Yifan Yang\textsuperscript{2,4,$\ddagger$}%
    \\[4pt]
    \textsuperscript{1}Atlas Lab \qquad
    \textsuperscript{2}School of Aerospace, BUAA \qquad
    \textsuperscript{3}School of Software, BUAA \\
    \textsuperscript{4}State Key Laboratory of HERATT \qquad
    \textsuperscript{5}Key Laboratory of SDODS (MOE)\\[4pt]
    Project Github:~\href{https://wallelwan.github.io/EagleVision}{https://wallelwan.github.io/EagleVision}
}
\author{\authorBlock}

\begin{document}
\twocolumn[{%
  \renewcommand\twocolumn[1][]{#1}%
  \vspace{-12mm}
  \maketitle
  \vspace{-10mm}
  \begin{center}
    \captionsetup{type=figure}
    \includegraphics[width=\linewidth]{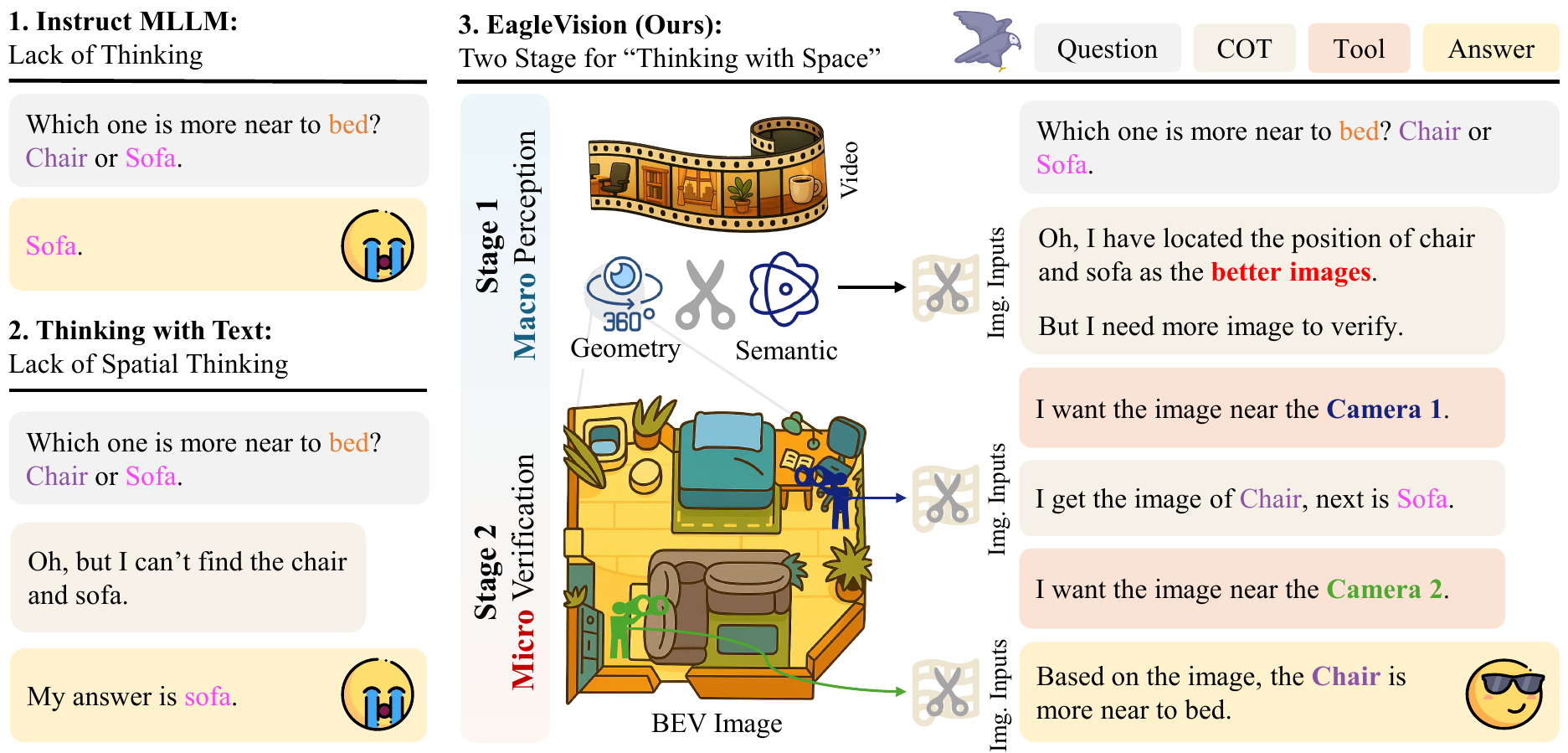}
    \vspace{-4mm}
    \captionof{figure}{\textbf{Overview of EagleVision.}
      Conventional MLLMs either answer spatial questions without reasoning (1) or reason in text only without accessing additional views (2). EagleVision introduces a dual-stage framework: (1)~\emph{Macro perception} selects a compact set of keyframes by jointly optimizing semantic relevance and geometric (viewpoint) diversity under a token budget. (2)~\emph{Micro verification} performs iterative spatial Chain-of-Thought (CoT) with active view selection---the model reasons in text, requests new frames by querying poses on a Bird's-Eye-View (BEV) map, and refines its answer through a closed-loop \emph{hypothesize--look--verify} cycle.}%
    \label{fig:teaser}
  \end{center}
}]

\renewcommand{\thefootnote}{}%
\footnotetext{\hspace{-1.8em}\textsuperscript{$\dagger$}Co-first Author. \quad
\textsuperscript{$\clubsuit$}Work done during internship at Atlas Lab. \quad
\textsuperscript{$\diamondsuit$}Project Lead. \quad
\textsuperscript{$\ddagger$}Corresponding author.}%
\addtocounter{footnote}{-1}%

\begin{abstract}
Video-based spatial reasoning—such as estimating distances, judging directions, or understanding layouts from multiple views—requires selecting informative frames and, when needed, actively seeking additional viewpoints during inference. Existing multimodal large language models (MLLMs) consume a fixed set of uniformly sampled frames and cannot request new views once reasoning begins, often missing the geometric cues necessary for reliable spatial judgments. We present \textbf{EagleVision}, a dual-stage framework that combines geometry-aware frame selection with active, Bird's-Eye-View (BEV)-grounded reasoning. In the first stage (\emph{macro perception}), a semantics--perspective-fusion determinantal point process (SPF-DPP) selects a compact set of keyframes that jointly maximize semantic relevance and viewpoint diversity under a fixed token budget. In the second stage (\emph{micro verification}), the model performs iterative spatial Chain-of-Thought: at each step it can either reason in text or predict a pose on the BEV plane to retrieve the nearest real frame, forming a closed-loop \emph{hypothesize--look--verify} cycle. The querying policy is trained purely via reinforcement learning with a spatial grounding reward, requiring no human-annotated reasoning traces. On VSI-Bench and SQA3D, EagleVision achieves state-of-the-art performance among open-source vision--language models.
\end{abstract}
\vspace{-6mm}

\section{Introduction}

Spatial reasoning over videos—estimating distances between objects, judging relative directions, or understanding room layouts from a sequence of frames—is a core capability for multimodal understanding~\cite{chen2024spatialvlm,cheng2024spatialrgpt,driess2023palm,habitat}, autonomous driving applications~\cite{zhang2026minddriver, yuan2025unimapgen,liang2025persistent,zeng2024priordrive}, and many downstream tasks~\cite{zhang2024sparse, zhang2026mgfnet, wan2025online, wan2025sp2t, zhang2024p2ftrack}.

Unlike single-image question answering, spatial reasoning require integrating geometric cues across multiple viewpoints: the model must decide \emph{which frames} provide the spatial information needed to answer a query, and sometimes the answer depends on views that were not initially selected. 
In a long video, however, spatial information is unevenly distributed—a small subset of frames may capture the critical viewpoint changes, while the majority are redundant.
Current multimodal large language models (MLLMs)~\cite{bai2023qwenvl,hurst2024gpt} are constrained by fixed token budgets and typically consume a set of uniformly sampled frames, which risks missing exactly those key viewpoints that carry the geometric parallax needed for reliable 3D inference~\cite{yang2024thinking}.
Worse still, once the initial frames are selected, existing models have no mechanism to request additional views during reasoning when the available evidence turns out to be insufficient.

Recent work on spatial intelligence in MLLMs has explored two main directions. 
The first augments 2D reasoning pipelines with 3D features such as point clouds or depth maps~\cite{chen2024spatialvlm,cheng2024spatialrgpt,hong20233d,qi2024shapellm,man2024situational,li2025amap, shan2025stability, hao2024mapdistill}, but these methods lack a mechanism for the model to actively seek additional views when its initial evidence is insufficient. 
The second direction couples MLLMs with 3D reconstruction modules~\cite{zhu2024llava,zheng2024video,fan2025vlm} that produce dense geometric representations; however, the reconstruction is typically a black-box preprocessing step whose outputs cannot be iteratively refined during reasoning.
Meanwhile, agent-based Chain-of-Thought (CoT) frameworks such as ChatGPT-o3~\cite{o3} and DeepEyes~\cite{zheng2025deepeyes} have shown that interleaving textual reasoning with active acquisition of visual evidence can improve image-level understanding.
However, these methods operate on single images (e.g., cropping or zooming) and do not address the multi-view, geometry-aware evidence gathering that spatial tasks demand.

Extending this ``thinking with images'' paradigm to video-based spatial reasoning introduces three concrete challenges.
First, MLLMs operate under strict token budgets, so the initial frame selection must balance \emph{semantic relevance} to the query with \emph{geometric diversity} of camera viewpoints—naive uniform sampling often fails at both.
Second, spatial hypotheses (e.g., ``object A is 2 meters left of object B'') need to be checked against specific views in a shared coordinate system, requiring a mechanism that maps abstract spatial queries to concrete video frames.
Third, collecting human annotations for multi-step spatial reasoning trajectories is impractical, so the model must learn when and where to query new views from answer-level supervision alone.

We propose \textbf{EagleVision}, a framework that addresses these challenges through two complementary stages.
In the first stage, \textbf{macro perception}, we select a compact set of keyframes that are both relevant to the query and geometrically diverse. 
Concretely, we build a Bird's-Eye View (BEV) map of the scene from an off-the-shelf SLAM system and use a \textbf{Semantics--Perspective-Fusion Determinantal Point Process (SPF-DPP)} to jointly optimize for semantic importance and SE(3) viewpoint diversity under a fixed token budget. 

In the second stage, \textbf{micro verification}, the model performs iterative spatial CoT with active view selection. 
At each reasoning step, the model can either produce text or request a new view by predicting a pose on the BEV plane.
The system retrieves the nearest real frame to the predicted pose via a scale-aware distance metric and appends it to the model's context, forming a closed-loop ``\emph{hypothesize $\to$ look $\to$ verify}'' cycle. 
We call this \textbf{BEV-grounded pose querying}. The entire querying policy is trained via Group Relative Policy Optimization (GRPO)~\cite{guo2025deepseek} with a spatial grounding reward that penalizes queries directed at regions lacking camera coverage, without requiring any human-annotated CoT supervision.

In summary, our contributions are:
\textbf{(1)}~We propose EagleVision, a dual-stage framework that combines geometry-aware frame selection with active BEV-grounded reasoning for video spatial understanding.
\textbf{(2)}~For macro perception, we introduce SPF-DPP, a sampling strategy that jointly models semantic relevance and viewpoint diversity to select informative keyframes under token constraints.
\textbf{(3)}~For micro verification, we formulate spatial CoT as BEV-grounded pose querying trained purely with reinforcement learning and a spatial grounding reward, enabling the model to actively acquire new views during reasoning without CoT annotations.
\textbf{(4)}~EagleVision achieves state-of-the-art performance among open-source VLMs on both VSI-Bench~\cite{yang2024thinking} and SQA3D~\cite{ma2022sqa3d}, demonstrating strong cross-benchmark generalization.
\begin{figure*}[t]
    \centering
    \includegraphics[width=0.75\linewidth]{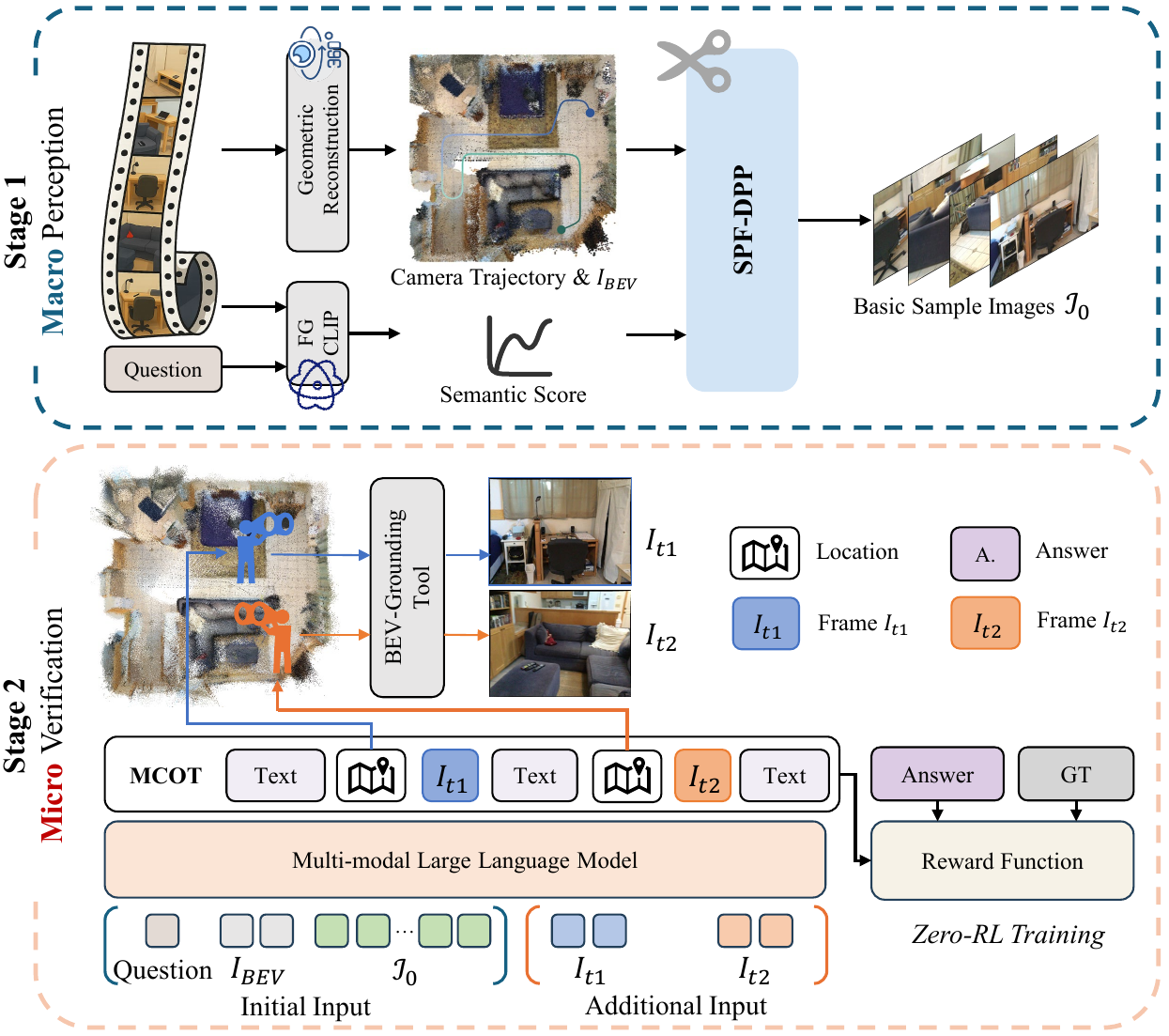}
    \vspace{-2mm}
    \caption{\textbf{Framework of EagleVision.} The framework operates in two stages: (i)~Macro perception selects spatially informative keyframes under a token budget by jointly optimizing semantic relevance and viewpoint diversity. (ii)~Micro verification performs iterative spatial CoT with active BEV-grounded pose querying to refine spatial understanding.}
    \vspace{-2mm}
    \label{fig:overview}
\end{figure*}

\section{Related Work}

\paragraph{Spatial-Intelligence MLLMs.}
Recent efforts to equip MLLMs with spatial reasoning follow three main directions:
(i)~incorporating explicit 3D representations such as point clouds or voxels into the language model~\cite{qi2024shapellm,man2024situational,chen2024spatialvlm,deng20253d,hong20233d,zhang2024sparse,wan2025sp2t};
(ii)~integrating 3D reconstruction modules as geometric priors~\cite{fan2025vlm,wang2024dust3r,zhu2024llava,wang2025ross3d,cheng2024spatialrgpt,zheng2024video}, with recent work such as Spatial-MLLM~\cite{wu2025spatial} fusing a VGGT-based spatial encoder with a 2D visual encoder;
and (iii)~replacing explicit 3D inputs with structured 2D projections, as in Struct2D~\cite{zhu2025struct2d} and SpatialMind~\cite{zhang2025spatial}.
All three directions improve the spatial information \emph{fed to} the model, but none allow the model to \emph{actively request additional views} during reasoning---the core capability introduced by EagleVision.

\paragraph{Interleaved-Modal CoT.}
Interleaved-modal CoT~\cite{gao2025interleaved} alternates visual evidence and textual reasoning within a single rollout. ChatGPT-o3~\cite{o3} and DeepEyes~\cite{zheng2025deepeyes} demonstrate that such tool-augmented reasoning can be learned via reinforcement learning~\cite{sanwal2025layered,hong2025deepeyesv2,yu2025sta}, but their tools operate on single images (cropping, zooming) rather than across views in a shared 3D coordinate system.
EagleVision extends this paradigm to spatial reasoning: the model's tool is a BEV-grounded pose query that retrieves real video frames, trained with a spatial grounding reward.
\section{Method}

\subsection{Overall Architecture}

\paragraph{Framework Overview.}
As illustrated in Fig.~\ref{fig:overview}, EagleVision decouples video-based spatial reasoning into two stages: (i)~\emph{macro perception}, which selects a compact yet spatially informative set of keyframes under a token budget, and (ii)~\emph{micro verification}, which performs iterative multimodal Chain-of-Thought (MCoT) reasoning with active view selection via BEV-grounded pose querying.

Given a video and a textual query, EagleVision first runs a frozen, off-the-shelf SLAM system to recover per-frame camera poses and depth maps. This reconstruction is a \textbf{one-time offline} preprocessing step that builds a reusable spatial index (BEV map and pose database); all subsequent reasoning is driven by a lightweight 2D VLM with simple nearest-neighbor lookups, incurring negligible per-query cost. This contrasts with 3D-capable LLMs that must re-invoke heavy 3D encoders at every inference call, making EagleVision's amortized cost lower in multi-query scenarios over the same scene. The framework is also \textbf{backend-agnostic}: the SLAM system can be replaced by other pose estimation methods (e.g., VGGT~\cite{wang2025vggt}) without degradation (see Sec.~\ref{sec:exp}).

\begin{figure*}[t]
    \centering
    \includegraphics[width=0.75\linewidth]{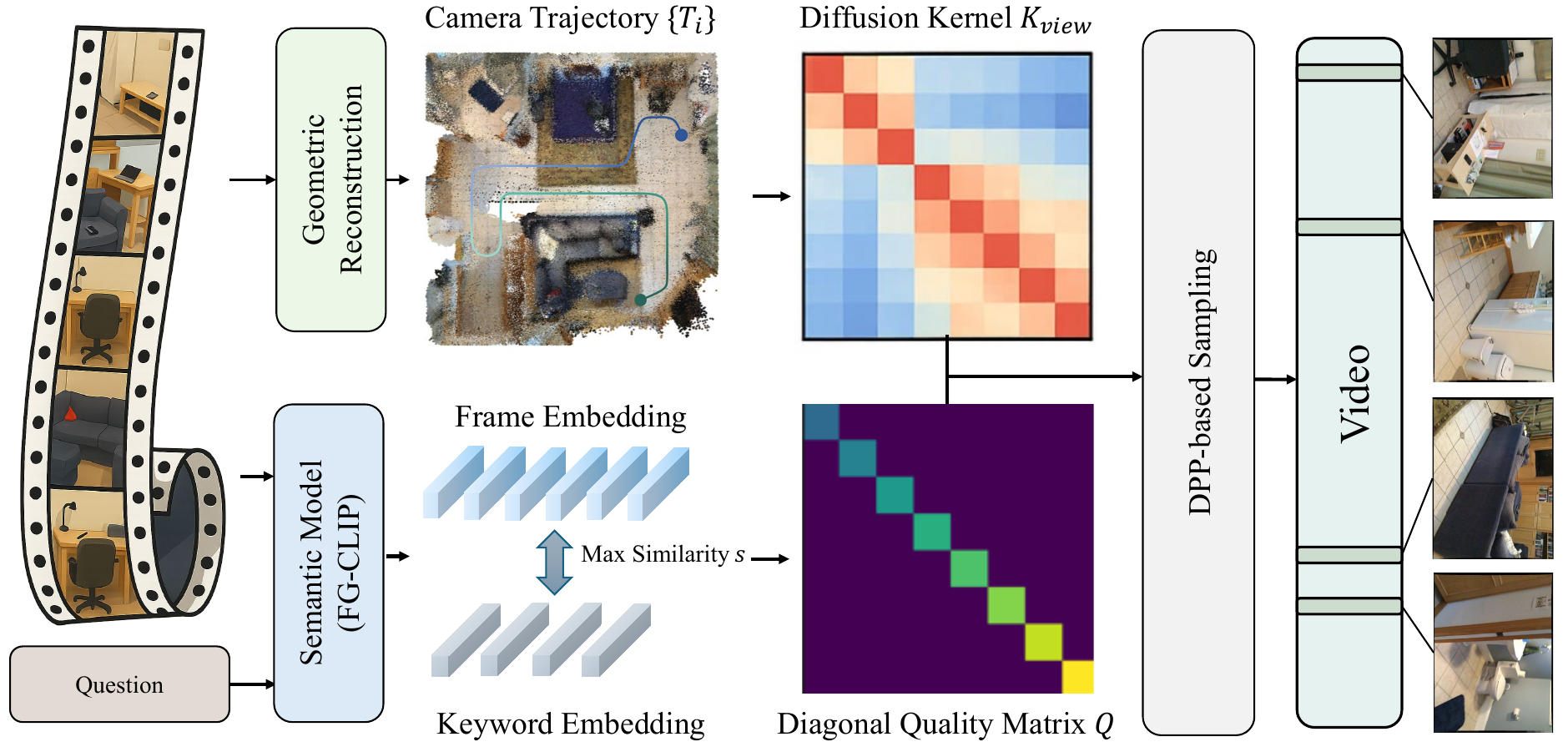}
    \vspace{-2mm}
    \caption{\textbf{Macro Perception.} Given an input video and a question, we first reconstruct the scene geometry and camera trajectory in $\mathrm{SE}(3)$ to build a sparse pose graph, whose heat-kernel diffusion yields the viewpoint kernel $K_{\mathrm{view}}$. In parallel, FG-CLIP~\cite{xie2025fg} computes frame--keyword similarities that are temperature-calibrated into semantic scores and encoded in a diagonal quality matrix $Q$. The resulting DPP L-ensemble $L_{\mathrm{dpp}} = Q K_{\mathrm{view}} Q$ is then used to greedily select $k$ frames.}
    \vspace{-2mm}
    \label{fig:dpp}
\end{figure*}

\paragraph{Macro Perception.}
The recovered poses and depths are projected onto a BEV plane to obtain geometry-aligned representations. In parallel, FG-CLIP~\cite{xie2025fg} computes the semantic relevance between each frame and the textual query.
These cues are jointly fed into SPF-DPP (Sec.~\ref{sec:spf-dpp}), which selects $k$ frames balancing viewpoint diversity and semantic importance. The selected keyframes, together with BEV projections, provide a global spatial prior for the reasoning stage.

\paragraph{Micro Verification.}
Starting from the selected keyframes and the BEV image, the model performs iterative spatial reasoning (Sec.~\ref{sec:agent_rl}). At each step, the model can either generate text, predict a pose on the BEV plane to retrieve a new frame, or terminate with an answer---forming a closed-loop \emph{hypothesize--look--verify} cycle. The querying policy is trained via reinforcement learning with a spatial grounding reward, without any human-annotated CoT supervision.

\subsection{SPF-DPP for Macro Perception}
\label{sec:spf-dpp}

Determinantal Point Processes (DPPs)~\cite{kulesza2012dpp} naturally model the trade-off between item quality and pairwise diversity in subset selection, which directly corresponds to our need for selecting frames that are both semantically relevant to the query and geometrically diverse in viewpoint.

\paragraph{Task Definition.}
As shown in Fig.~\ref{fig:dpp}, given a video with $N$ frames, each frame $i$ has a camera pose $T_i=(R_i,\mathbf{t}_i)\!\in\!\mathrm{SE}(3)$ and a semantic score $s_i\!\in\![0,1]$ computed w.r.t.\ the query.
Our goal is to select a fixed-size subset $X\subseteq\{1,\dots,N\}$ with $|X|=k$ that is diverse in viewpoint while semantically representative.

\paragraph{Pose Distance and Sparse Viewpoint Graph.}
We measure pairwise frame distance in $\mathrm{SE}(3)$ by combining translation and rotation:
{\footnotesize
\begin{align*}
d_{ij}^2 \;=\;& 
\underbrace{\big\|\mathbf{t}_i-\mathbf{t}_j\big\|^2 / \sigma_t^2}_{\text{translation}}
\;+\;
\underbrace{\beta^2\,\theta(R_i,R_j)^2}_{\text{rotation}},\\
\theta(R_i,R_j) \;=\;&
\arccos\!\Big(\tfrac{\mathrm{tr}(R_i^\top R_j)-1}{2}\Big).
\end{align*}
}

\noindent where $\sigma_t$ normalizes scene scale and $\beta$ balances rotation against translation.
Since the BEV projection applies a fixed metric normalization via the oriented bounding box (see supplementary material), the spatial scale is deterministic for each scene, and $\sigma_t$ can be set as a fixed constant without requiring adaptive search (e.g., bisection).
We convert distances to affinities $w_{ij}=\exp(-\frac{1}{2}d_{ij}^2)$ and retain only edges within a temporal window of bandwidth $b$:
\[
W_{ij} \;=\; 
\begin{cases}
w_{ij}, & \text{if } |i-j|\le b,\\
0, & \text{otherwise},
\end{cases}
\]
yielding a sparse, symmetric adjacency matrix $W$ that keeps the subsequent computation efficient.

\begin{figure*}[t]
    \centering
    \includegraphics[width=0.65\linewidth]{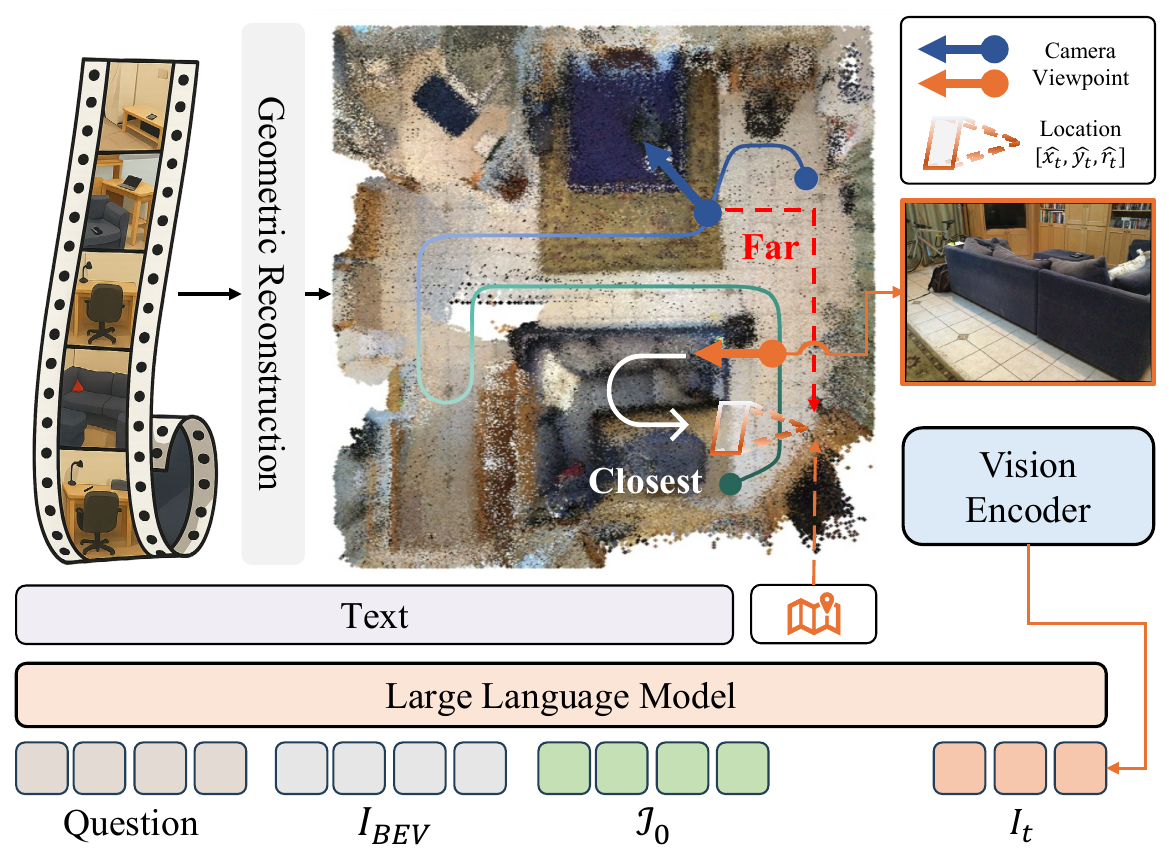}
    \vspace{-2mm}
    \caption{\textbf{Micro Verification via Spatial MCoT.} 
Given the selected keyframes and a BEV image, the model reasons in text and actively queries poses on the BEV plane to retrieve additional video frames when spatial evidence is insufficient, iteratively refining its answer.}
    \vspace{-2mm}
    \label{fig:mcot}
\end{figure*}

\paragraph{Diffusion Viewpoint Kernel.}
To propagate local viewpoint affinities into global geometric relations, we apply a heat-kernel diffusion on the graph Laplacian:
\[
K_{\mathrm{view}} \;=\; \exp\!\big(-\tau\,\mathcal{L}\big),
\quad
\mathcal{L}=I-D^{-1/2}WD^{-1/2},
\]
where $\tau>0$ is the diffusion scale and $D$ is the degree matrix. The resulting $K_{\mathrm{view}}$ is positive semidefinite (PSD) and encodes how geometrically similar any two frames are from a viewpoint coverage perspective.

\paragraph{Semantic Score and Quality Modulation.}
Raw semantic scores $s_i$ are calibrated via temperature-scaled softmax to mitigate scale effects:
\[
\tilde{s}_i \;=\; \mathrm{norm}_{0\text{-}1}\!\Big(\frac{\exp(s_i/T)}{\sum_j \exp(s_j/T)}\Big), \quad T>0.
\]
We define a diagonal quality matrix
$Q =\mathrm{diag}(q_1,\ldots,q_N)$ with
$q_i = (1{-}\alpha)+\alpha\,\tilde{s}_i$, $\alpha\in[0,1]$,
which biases selection toward query-relevant frames while keeping a nonzero floor $(1{-}\alpha)$ to preserve diversity.

\paragraph{L-ensemble Kernel and Fixed-Size Selection.}
The DPP L-ensemble kernel is $L_{\mathrm{dpp}} = Q\,K_{\mathrm{view}}\,Q$,
which is PSD because $K_{\mathrm{view}}$ is PSD and $Q$ is diagonal nonnegative.
For a fixed budget $k$, we solve
\[
\max_{X\subseteq[N],\,|X|=k}\;\; \log\det\!\big(L_{\mathrm{dpp},X}\big)
\]
via greedy MAP with rank-one Cholesky updates~\cite{kulesza2012dpp}, which gives a $(1{-}1/e)$-approximation. Further details on hyper-parameters are in the supplementary material.

\subsection{Spatial MCoT for Micro Verification}
\label{sec:agent_rl}

We refer to the iterative reasoning process of the micro verification stage as \emph{Spatial Multimodal Chain-of-Thought} (Spatial MCoT). At each reasoning step $t$, the model maintains a state consisting of all previously generated text tokens $\mathbf{X}_{\le t}$, all retrieved visual observations $\mathbf{I}_{\le t}$, and a BEV pose buffer $\mathcal{B}$ recording past queries. The model then selects one of three actions: (i)~emit a text token to continue reasoning, (ii)~issue a pose query on the BEV plane to retrieve a new frame, or (iii)~terminate with the current answer.

\begin{table*}[ht!]
    \captionsetup{type=table}
    \centering
    \begin{minipage}{0.83\textwidth}
    \centering
    \fontsize{4.6pt}{4.4pt}\selectfont
    \setlength\tabcolsep{3pt}
    \renewcommand{\arraystretch}{1.2}
    \resizebox{\textwidth}{!}{
    \begin{tabular}{r|ccc|cccccccc}
& & & &
\rotatebox{75}{Obj. Count} &
\rotatebox{75}{Abs. Dist.} &
\rotatebox{75}{Obj. Size} &
\rotatebox{75}{Room Size} &
\rotatebox{75}{Rel. Dist.} &
\rotatebox{75}{Rel. Dir.} &
\rotatebox{75}{Route Plan} &
\rotatebox{75}{Appr. Order} \\
Methods & Rank & SQA-3D & VSI-Bench &
\multicolumn{4}{c}{\cellcolor{orange!10}Numerical Answer} &
\multicolumn{4}{c}{\cellcolor{yellow!10}Multiple-Choice Answer} \\
\hline
\rowcolor{navyblue!5}
\multicolumn{12}{l}{\textit{Baseline}} \\
Chance Level (Random) & - & - & - & - & - & - & - & 25.0 & 36.1 & 28.3 & 25.0 \\
Chance Level (Frequency) & - & - & 34.0 & 62.1 & 32.0 & 29.9 & 33.1 & 25.1 & 47.9 & 28.4 & 25.2 \\
\hline
\rowcolor{navyblue!5}
\multicolumn{12}{l}{\textit{VSI-Bench Perf. (\dag = Tiny Set)}}~\cite{yang2024thinking} \\
\dag Human Level & - & - & 79.2 & 94.3 & 47.0 & 60.4 & 45.9 & 94.7 & 95.8 & 95.8 & 100.0 \\
\dag Gemini-1.5 Flash~\cite{anil2024gemini} & - & - & 45.7 & 50.8 & 33.6 & 56.5 & 45.2 & 48.0 & 39.8 & 32.7 & 59.2 \\
\dag Gemini-1.5 Pro~\cite{anil2024gemini} & - & - & 48.8 & 49.6 & 28.8 & 58.6 & 49.4 & 46.0 & 48.1 & 42.0 & 68.0 \\
\dag Gemini-2.0 Flash~\cite{anil2024gemini} & - & - & 45.4 & 52.4 & 30.6 & 66.7 & 31.8 & 56.0 & 46.3 & 24.5 & 55.1 \\
\hline
\rowcolor{navyblue!5}
\multicolumn{12}{l}{\textit{Proprietary Models (API)}} \\
GPT-4o~\cite{hurst2024gpt} & \cellcolor{oai-green-200}{3} & - & 34.0 & 46.2 & 5.3 & 43.8 & 38.2 & 37.0 & 41.3 & 31.5 & 28.5 \\
Gemini-1.5 Flash~\cite{anil2024gemini} & \cellcolor{oai-green-400}{2} & - & 42.1 & 49.8 & 30.8 & 53.5 & 54.4 & 37.7 & 41.0 & 31.5 & 37.8 \\
Gemini-1.5 Pro~\cite{anil2024gemini} & \cellcolor{oai-green-600}{1} & - & 45.4 & 56.2 & 30.9 & 64.1 & 43.6 & 51.3 & 46.3 & 36.0 & 34.6 \\
\hline
\rowcolor{navyblue!5}
\multicolumn{12}{l}{\textit{Open-sourced VLMs}} \\
InternVL2-2B~\cite{chen2024internvl} & 14 & - & 27.4 & 21.8 & 24.9 & 22.0 & 35.0 & 33.8 & 44.2 & 30.5 & 7.1 \\
LLaVA-OneVision-0.5B~\cite{li2024llava} & 13 & - & 28.0 & 46.1 & 28.4 & 15.4 & 28.3 & 28.9 & 36.9 & 34.5 & 5.8 \\
LongVA-7B~\cite{zhang2024long} & 12 & - & 29.2 & 38.0 & 16.6 & 38.9 & 22.2 & 33.1 & 43.3 & 25.4 & 15.7 \\
LLaVA-OneVision-7B~\cite{li2024llava} & 11 & - & 32.4 & 47.7 & 20.2 & 47.4 & 12.3 & 42.5 & 35.2 & 29.4 & 24.4 \\
InternVL2-8B~\cite{chen2024internvl} & 10 & - & 34.6 & 23.1 & 28.7 & 48.2 & 39.8 & 36.7 & 30.7 & 29.9 & 39.6 \\
LLaVA-NeXT-Video-7B~\cite{li2024llavanext} & 9 & - & 35.6 & 48.5 & 14.0 & 47.8 & 24.2 & 43.5 & 42.4 & 34.0 & 30.6 \\
Qwen2.5-VL-7B~\cite{bai2023qwenvl} & 7 & - & 35.8 & -- & -- & -- & -- & -- & -- & -- & -- \\
Struct2D~\cite{zhu2025struct2d} & 6 & 58.5 & 43.6 &47.1 & 35.1 & 57.1 &48.9 & 35.1 & 45.9 & 35.8 & -- \\
Spatial-Mind~\cite{zhang2025spatial} & 5 & 46.3 & 47.1 & 48.6 & 34.4 & 68.9 & 54.7 & 53.4 & 43.9 & 30.1 & 42.7\\
Spatial-MLLM~\cite{wu2025spatial} & 4 & 55.9 & 48.4 & 65.3 & 34.8 & 63.1 & 45.1 & 41.3 & 46.2 & 33.5 & 46.3 \\
Qwen3-VL-8B~\cite{qwen3} & \cellcolor{oai-green-200}{3} & - & 59.4 & -- & -- & -- & -- & -- & -- & -- & -- \\
VLM-3R-7B~\cite{fan2025vlm} & \cellcolor{oai-green-400}{2} & - & 60.9 & 70.2 & 49.4 & 69.2 & 67.1 & 65.4 & 80.5 & 45.4 & 40.1 \\
\hline
\textbf{EagleVision (Ours)} &
\cellcolor{oai-green-600}{1} & 60.3 & 63.5 & 74.9 & 41.3 & 72.3 & 69.4 & 70.1 & 84.7 & 46.3 & 49.3 \\
\hline
\end{tabular}

    }
    \end{minipage}
    \vspace{-0.1cm}
    \caption{\textbf{Evaluations on SQA-3D (Val) and VSI-Bench.} EagleVision ranks first among open-source VLMs, showcasing the effectiveness of our dual-stage framework. \textsuperscript{\dag}Results on the VSI-Bench tiny set are presented following the setup in~\cite{yang2024thinking}.}
    \label{tab:vsibench}
    \vspace{-3mm}
\end{table*}

\paragraph{BEV-Grounded Pose Querying.}
When the model issues a query, it predicts a 2D BEV pose 
$\hat{p}_t=(\hat{x}_t,\hat{y}_t,\hat{r}_t)$, 
where $(\hat{x}_t,\hat{y}_t)$ are BEV coordinates and 
$\hat{r}_t$ is the orientation angle.
The system scores each video frame $j$ (with stored BEV position $(x_j,y_j)$ and orientation $r_j$) by spatial similarity:
\[
s_{tj}=\exp\!\Big[-\tfrac{1}{2}\big(\|(\hat{x}_t,\hat{y}_t)-(x_j,y_j)\|^2/\sigma_p^2 
+ \beta^2(\hat{r}_t-r_j)^2\big)\Big],
\]
where $\sigma_p$ controls the spatial scale and $\beta$ balances rotation sensitivity.
Intuitively, $s_{tj}$ measures how well frame $j$ matches the viewpoint the model is requesting.
The frame with the highest similarity $j^\star = \arg\max_j s_{tj}$ is retrieved if $s^{(t)}_{\max}=s_{t j^\star}\ge\tau_s$; otherwise, an error prompt is returned (indicating no frame covers the queried pose). The retrieved frame is appended to $\mathbf{I}_{\le t}$, and the total number of queries is capped at $T_{\max}$.

\paragraph{Reward Function.}
The trajectory reward consists of two groups of terms:
{\small
\[
R(\tau)\;=\;\underbrace{R_{\text{acc}}(\tau)+R_{\text{format}}(\tau)+\lambda_{\text{tool}}\,R_{\text{tool}}(\tau)}_{\text{task rewards (following DeepEyes~\cite{zheng2025deepeyes})}}
\;+\;\underbrace{\lambda_{\text{spatial}}\,R_{\text{spatial}}(\tau)}_{\text{spatial grounding (ours)}}.
\]
}
The first group is inherited from DeepEyes~\cite{zheng2025deepeyes}:
$R_{\text{acc}}$ rewards answer correctness,
$R_{\text{format}}$ encourages well-formed output structure, and
$R_{\text{tool}}$ incentivizes proper tool invocation.
We refer readers to~\cite{zheng2025deepeyes} for their precise definitions.
 
These task-level rewards alone do not prevent the model from querying geometrically invalid viewpoints---poses that fall outside the camera trajectory and therefore cannot be matched to any real frame.
To address this, we introduce a \textbf{spatial grounding reward} $R_{\text{spatial}}$ that penalizes trajectories containing any query whose best retrieval similarity falls below the coverage threshold~$\tau_s$:
\[
R_{\text{spatial}}(\tau)=
\begin{cases}
-1, & \text{if }\exists\,t\in\mathcal{C}(\tau)\;\text{s.t.}\; s_{\max}^{(t)} < \tau_s,\\[3pt]
\phantom{-}0, & \text{otherwise},
\end{cases}
\]
where $\mathcal{C}(\tau)$ denotes the set of query steps in trajectory~$\tau$ and $s_{\max}^{(t)}=\max_j s_{tj}$ is the best frame--query similarity at step~$t$ (defined in the preceding paragraph).
The penalty magnitude is controlled solely by $\lambda_{\text{spatial}}$; we set the per-term penalty to $-1$ for simplicity.
Intuitively, this reward teaches the model to \emph{query only within well-covered regions} of the scene, avoiding hallucinated viewpoints that would return uninformative error prompts.

\paragraph{Optimization.}
We train the querying policy using Group Relative Policy Optimization (GRPO)~\cite{guo2025deepseek}, which estimates advantages within prompt-level trajectory groups to reduce variance, with a frozen reference policy for KL regularization.
Following DeepSeek-R1~\cite{guo2025deepseek}, we train \textbf{directly with RL without a supervised fine-tuning (SFT) cold start}: the model learns spatial querying purely from the reward signal, without human-annotated reasoning traces.
Only model-generated text tokens contribute to the reward; retrieved observations and tool metadata are masked out.
\section{Experiment}
\label{sec:exp}

\subsection{Implementation Details}

\paragraph{Baselines and Benchmarks.}
We compare EagleVision with four categories of baselines: (1)~randomized heuristics (random and frequency-based), (2)~proprietary models (e.g., Gemini-1.5~\cite{anil2024gemini}, Gemini-2.0~\cite{anil2024gemini}), (3)~open-source VLMs (e.g., Qwen3-VL-8B~\cite{qwen3}), and (4)~task-specific spatial reasoning methods (e.g., VLM-3R-7B~\cite{fan2025vlm}, Struct2D~\cite{zhu2025struct2d}, Spatial-MLLM~\cite{wu2025spatial}, SpatialMind~\cite{zhang2025spatial}).
We evaluate on two benchmarks: \textbf{VSI-Bench}~\cite{yang2024thinking}, which assesses 3D spatial understanding across configuration and measurement tasks, and \textbf{SQA3D}~\cite{ma2022sqa3d}, which tests situated question answering in 3D scenes. All models are tested with identical video input and question templates.

\paragraph{Training Details.}
We fine-tune Qwen3-VL-8B~\cite{qwen3} using GRPO~\cite{guo2025deepseek} for 80 cycles on H20 GPUs. Each batch contains 256 prompts with 16 rollouts per prompt and a maximum of six tool-call steps. The KL coefficient is set to 0.0, and the response length is capped at 20,480 tokens.
The training data comes from the training split of VLM-3R~\cite{fan2025vlm} (and SQA3D~\cite{ma2022sqa3d} for the cross-benchmark evaluation), with no data leakage to the test sets.

\subsection{Main Results}

\paragraph{VSI-Bench.}
As shown in Tab.~\ref{tab:vsibench}, EagleVision achieves the best average score (63.5) among all open-source VLMs. Compared to the base model Qwen3-VL-8B~\cite{qwen3} (59.4), EagleVision improves by +4.1 points, demonstrating the effectiveness of combining geometry-aware frame selection with active BEV-grounded reasoning. Compared to VLM-3R-7B~\cite{fan2025vlm} (60.9), which uses an additional 3D reconstruction encoder, EagleVision achieves a +2.6-point gain without modifying the model architecture or its visual input pipeline.

\paragraph{SQA3D.}
To evaluate cross-benchmark generalization, we apply EagleVision to SQA3D~\cite{ma2022sqa3d} and compare with recent spatial reasoning methods in Tab.~\ref{tab:vsibench}. EagleVision achieves 60.3\% (EM@1), outperforming Struct2D~\cite{zhu2025struct2d} (+1.8\%), Spatial-MLLM~\cite{wu2025spatial} (+4.4\%), and SpatialMind~\cite{zhang2025spatial} (+14.0\%). This confirms that the RL-trained spatial querying paradigm transfers effectively across benchmarks and consistently outperforms recent fully-supervised methods.

\begin{table}[t]
    \centering
    \footnotesize
    \begin{tabularx}{\linewidth}{>{\centering\arraybackslash}X >{\centering\arraybackslash}X >{\centering\arraybackslash}X | >{\centering\arraybackslash}X}
\toprule
MCoT & BEV & SPF-DPP & Avg. \\
\midrule
\multicolumn{3}{c|}{Baseline (Qwen3-VL-8B)} & 59.4 \\
\usym{2714} & & & 61.9 \\
\usym{2714} & \usym{2714} & & 62.7 \\
\usym{2714} & & \usym{2714} & 62.5 \\
\rowcolor{natureblue}
\usym{2714} & \usym{2714} & \usym{2714} & 63.5 \\
\bottomrule
\end{tabularx}
    \caption{\textbf{Ablation study of overall architecture.} MCoT means Spatial MCoT; BEV means BEV image input; SPF-DPP means geometry-aware frame selection.}\label{tab:ablation_structure}
    \vspace{-1mm}
\end{table}

\begin{table}[t]
    \centering
    \footnotesize
    \begin{tabularx}{\linewidth}{>{\centering\arraybackslash}X >{\centering\arraybackslash}X >{\centering\arraybackslash}X | >{\centering\arraybackslash}X}
\toprule
w/o DPP & Geometry & Semantic & Avg. \\
\midrule
\usym{2714} & & & 62.7 \\
& \usym{2714} & & 63.2 \\
& & \usym{2714} & 63.0 \\
\rowcolor{natureblue}
& \usym{2714} & \usym{2714} & 63.5 \\
\bottomrule
\end{tabularx}
    \caption{\textbf{Ablation study of SPF-DPP components} on the MCoT+BEV baseline.}\label{tab:ablation_dpp}
    \vspace{-4mm}
\end{table}

\begin{table}[t]
    \centering
    \footnotesize
    \begin{tabularx}{\linewidth}{>{\centering\arraybackslash}X >{\centering\arraybackslash}X >{\centering\arraybackslash}X >{\centering\arraybackslash}X | >{\centering\arraybackslash}X}
\toprule
Acc. & Format & Tool & Spatial & Avg. \\
\midrule
\usym{2714} & \usym{2714} & & & 61.1 \\
\usym{2714} & \usym{2714} & \usym{2714} & & 63.0 \\
\rowcolor{natureblue}
\usym{2714} & \usym{2714} & \usym{2714} & \usym{2714} & 63.5 \\
\bottomrule
\end{tabularx}
    \caption{\textbf{Ablation study of GRPO reward terms.} Acc. means accuracy reward.}\label{tab:ablation_reward}
    \vspace{-4mm}
\end{table}

\subsection{Ablation Study}

All ablation variants are trained on VLM-3R~\cite{fan2025vlm} and evaluated on VSI-Bench~\cite{yang2024thinking} under identical settings.

\paragraph{Overall Architecture.}
We ablate the three components of EagleVision on Qwen3-VL-8B (59.4 Avg), as shown in Tab.~\ref{tab:ablation_structure}. Adding Spatial MCoT alone improves to 61.9, showing that iterative multimodal reasoning enhances spatial awareness. Introducing BEV grounding further lifts performance to 62.7 (+3.3 over baseline), confirming the benefit of explicit geometric anchoring via active view selection. Adding SPF-DPP on top yields the best 63.5 (+0.8), demonstrating that geometry-aware frame selection further improves initial evidence quality. The three components are complementary: MCoT provides the reasoning structure, BEV grounding enables spatial verification, and SPF-DPP optimizes the starting evidence set.

\paragraph{SPF-DPP.}
Tab.~\ref{tab:ablation_dpp} ablates the two SPF-DPP components---geometry-aware diffusion and semantic modulation---on the MCoT+BEV baseline (62.7). Geometry alone reaches 63.2 (+0.5), improving viewpoint diversity via pose-aware similarity. Semantics alone gives 63.0 (+0.3), favoring task-relevant frames. Combining both achieves 63.5 (+0.8), confirming their complementarity.

\begin{table}[t]
    \centering
    \footnotesize
    \begin{tabularx}{\linewidth}{>{\centering\arraybackslash}p{1.2em} | >{\centering\arraybackslash}X >{\centering\arraybackslash}X >{\centering\arraybackslash}X >{\centering\arraybackslash}X | >{\centering\arraybackslash}X}
\toprule
No. & $\beta$ & $\tau$ & $\alpha$ & $b$ & Score \\
\midrule
1 & 1 & 2.0 & 0.5 & 24 & 62.8 \\
\rowcolor{natureblue}
2 & 2 & 2.0 & 0.5 & 24 & 63.5 \\
3 & 4 & 2.0 & 0.5 & 24 & 63.1 \\
\midrule
4 & 2 & 1.0 & 0.5 & 24 & 62.1 \\
5 & 2 & 1.5 & 0.5 & 24 & 62.9 \\
\rowcolor{natureblue}
6 & 2 & 2.0 & 0.5 & 24 & 63.5 \\
\midrule
7 & 2 & 2.0 & 0.3 & 24 & 62.9 \\
\rowcolor{natureblue}
8 & 2 & 2.0 & 0.5 & 24 & 63.5 \\
9 & 2 & 2.0 & 0.7 & 24 & 62.3 \\
\midrule
10 & 2 & 2.0 & 0.5 & 12 & 63.3 \\
\rowcolor{natureblue}
11 & 2 & 2.0 & 0.5 & 24 & 63.5 \\
12 & 2 & 2.0 & 0.5 & 36 & 63.5 \\
\bottomrule
\end{tabularx}
    \caption{\textbf{SPF-DPP hyper-parameter sensitivity.} Each parameter is varied while others are held at defaults ($\beta{=}2$, $\tau{=}2$, $\alpha{=}0.5$, $b{=}24$). Default settings are in the supplementary.}\label{tab:spf_hyper}
    \vspace{-4mm}
\end{table}

\paragraph{SPF-DPP Hyper-parameter Sensitivity.}
To address concerns about parameter sensitivity, we vary each of the four key hyper-parameters individually while fixing the rest at their defaults (Tab.~\ref{tab:spf_hyper}). The rotation weight $\beta$ and temporal bandwidth $b$ are largely robust: performance stays within 0.5 points across tested values. The diffusion scale $\tau$ and quality mixing weight $\alpha$ are more sensitive, yet SPF-DPP consistently outperforms naive time-based sampling across all tested configurations. This confirms that the gains stem from explicitly modeling diversity rather than from a narrow hyper-parameter sweet spot.

\paragraph{GRPO Reward.}
Tab.~\ref{tab:ablation_reward} ablates the reward terms. Accuracy + Format yields 61.1. Adding the Tool bonus boosts performance to 63.0 (+1.9), showing that outcome-conditioned encouragement of tool use is crucial. Adding the Spatial term reaches 63.5 (+0.5), by rewarding pose-consistent calls and penalizing invalid retrievals. Overall, the Tool reward provides the main gain, while the Spatial term further regularizes querying decisions.

\paragraph{Robustness to Pose Noise.}
To assess sensitivity to pose estimation errors, we inject Gaussian noise into the camera trajectory at three levels and also replace Vipe with a learning-based alternative (VGGT~\cite{wang2025vggt}). As shown in Tab.~\ref{tab:noise}, performance degrades gracefully: even with medium noise ($5\%$, $5^\circ$), accuracy drops by only 0.4\%. Replacing the backend with VGGT yields 63.2 ($-0.3$), confirming that the gains stem from the macro--micro framework rather than a specific SLAM system. This robustness arises because nearest-frame retrieval acts as a spatial buffer---even under coordinate drift, retrieved frames retain sufficient visual overlap for effective grounding.

\begin{table}[t]
    \centering
    \footnotesize
    \begin{tabularx}{\linewidth}{>{\raggedright\arraybackslash}X c >{\centering\arraybackslash}p{3em}}
\toprule
Setting / Noise Level & VSI-Bench & Drop \\
\midrule
Clean (Original) & 63.5 & -- \\
Mild ($2\%, 2^\circ$) & 63.3 & $-$0.2 \\
Medium ($5\%, 5^\circ$) & 63.1 & $-$0.4 \\
Heavy ($10\%, 10^\circ$) & 62.9 & $-$0.6 \\
\midrule
\textit{Vipe $\to$ VGGT} & 63.2 & $-$0.3 \\
\bottomrule
\end{tabularx}
    \caption{\textbf{Robustness under pose noise} and with VGGT backend.}
    \label{tab:noise}
    \vspace{-4mm}
\end{table}

\begin{figure}[t]
    \centering
    \includegraphics[width=\linewidth]{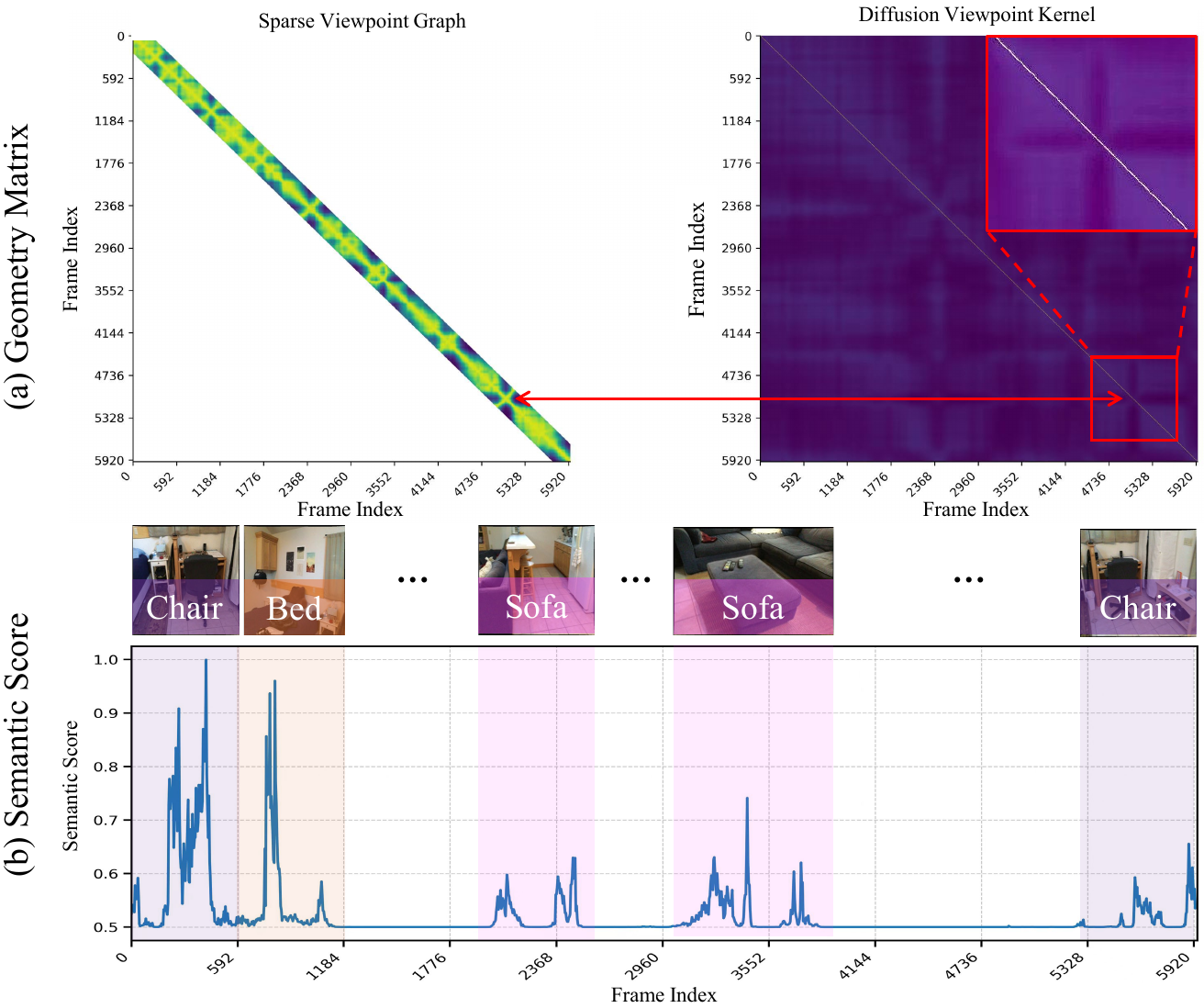}
    \vspace{-6mm}
    \caption{\textbf{Visualization of SPF-DPP} on ScanNet Scene$00\_00$. (a)~Sparse Viewpoint Graph vs.\ Diffusion Viewpoint Kernel. (b)~Semantic scores for keywords \emph{chair}, \emph{bed}, and \emph{sofa} with representative peak frames.}
    \vspace{-6mm}
    \label{fig:vis2}
\end{figure}

\begin{figure}[t]
    \centering
    \includegraphics[width=\linewidth]{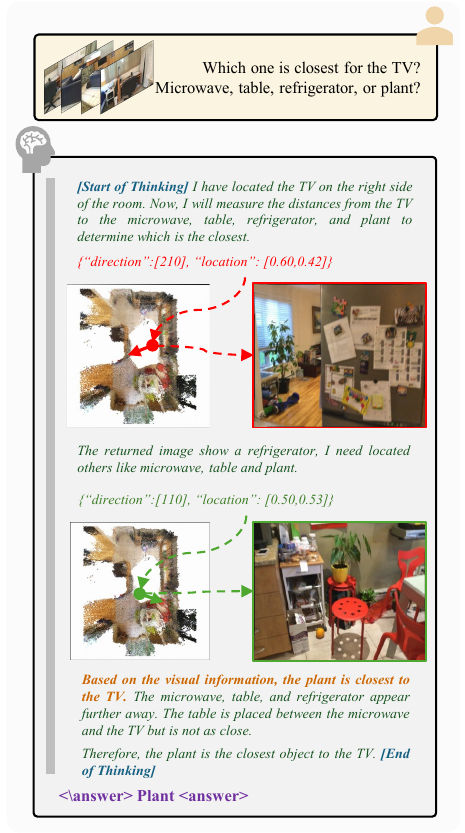}
    \vspace{-6mm}
    \caption{\textbf{Qualitative example.} The model iteratively queries BEV poses to locate each object and determine which is nearest to the TV.}
    \vspace{-6mm}
    \label{fig:vis}
\end{figure}

\subsection{Visualization}
\paragraph{SPF-DPP.}
Fig.~\ref{fig:vis2}(a) compares the Sparse Viewpoint Graph with the Diffusion Viewpoint Kernel. The sparse graph's affinities $W_{ij}$ decay with temporal distance, leaving many zero entries; the diffusion kernel propagates them globally, attenuating sharply where the camera moves fast and retaining similarity where it moves slowly. This lets SPF-DPP favor complementary viewpoints over redundant nearby frames.
Fig.~\ref{fig:vis2}(b) confirms that semantic scores peak only when query-relevant objects are visible, underscoring the need to combine geometric and semantic cues.
 
\paragraph{Model Response.}
Fig.~\ref{fig:vis} illustrates the full reasoning pipeline. Asked ``Which one is closest to the TV---microwave, table, refrigerator, or plant?'', the model anchors the TV and enters an iterative \emph{hypothesize--look--verify} loop. The first BEV query (${\sim}210^{\circ}$, position $(0.60,0.42)$) retrieves a close-up of the refrigerator; the second (${\sim}110^{\circ}$, $(0.50,0.53)$) reveals the plant. After each retrieval the model updates its spatial estimate rather than committing prematurely, ultimately concluding the plant is nearest. This highlights two strengths: the ability to \emph{actively} select informative viewpoints on demand, and the capacity to integrate evidence across retrievals into a coherent spatial judgment.
\section{Conclusion}

We present EagleVision, a dual-stage framework for video-based spatial reasoning. For macro perception, we propose SPF-DPP, which jointly optimizes semantic relevance and viewpoint diversity to select informative keyframes under a token budget. For micro verification, we formulate spatial Chain-of-Thought as BEV-grounded pose querying---enabling the model to actively request new views during reasoning---and train the querying policy purely via reinforcement learning with a spatial grounding reward, without any CoT supervision. Experiments on VSI-Bench and SQA3D demonstrate state-of-the-art performance among open-source VLMs.

\section*{Acknowledgments}
This work was supported by the National Natural Science Foundation of China (62433003) and the National Natural Science Foundation of China (62476017). Project GitHub: https://wallelwan.github.io/EagleVision

{
    \small
    \bibliographystyle{ieeenat_fullname}
    \bibliography{main}
}

\clearpage
\setcounter{page}{1}
\maketitlesupplementary
\appendix

\renewcommand\thetable{A\arabic{table}}
\setcounter{table}{0}
\renewcommand\thefigure{A\arabic{figure}}
\setcounter{figure}{0}

\begin{strip}
    \centering
    \includegraphics[width=\linewidth]{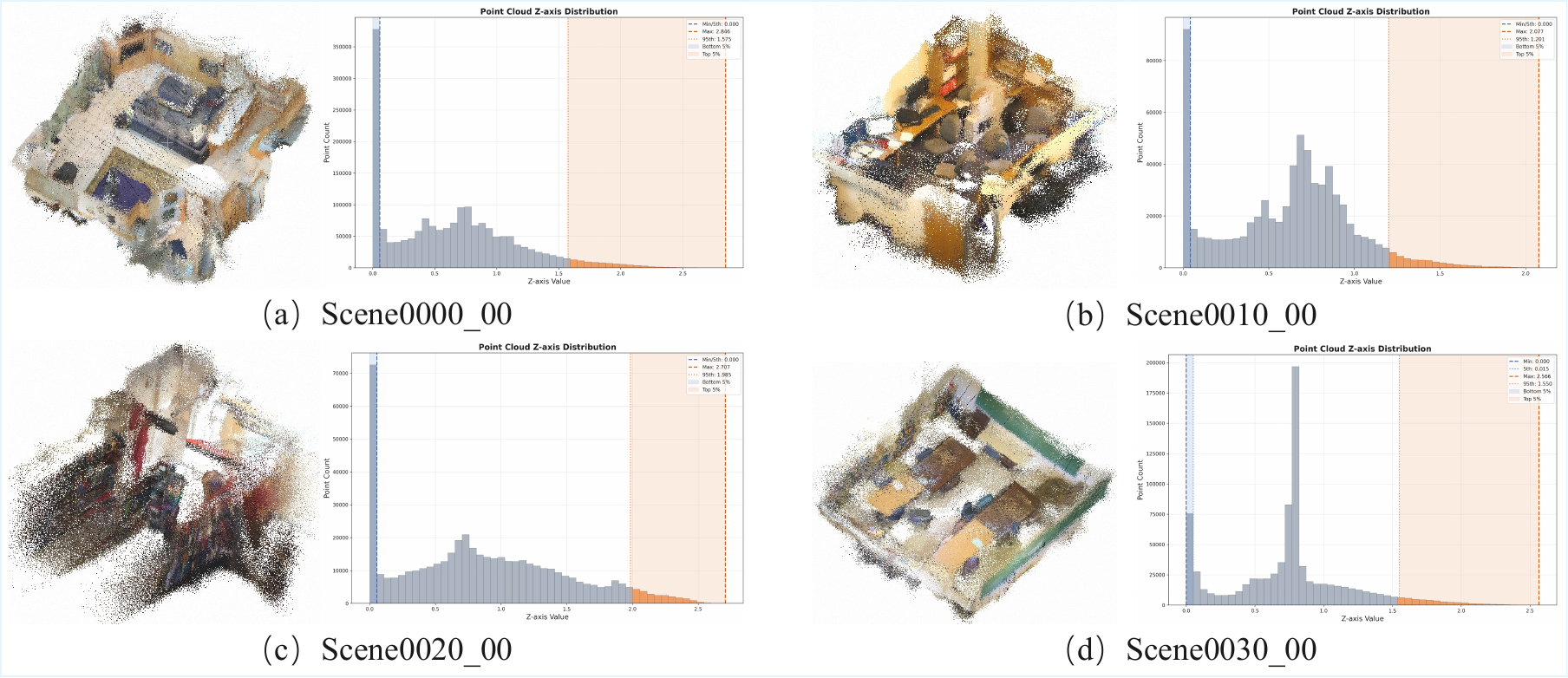}
    \captionof{figure}{Visualization of point cloud and z-axis histogram.} 
    \label{fig:z_vis}
\end{strip}

\section{3D Reconstruction Pre-processing Pipeline}
\label{sec:reconstruction}
We build a lightweight 3D reconstruction pre-processing pipeline that converts raw videos into a ground-aligned bird's-eye-view (BEV) representation. It consists of four steps: multi-view 3D reconstruction, point cloud bounding-box normalization, ground-aware plane estimation, and BEV generation.

\subsection{Multi-view 3D Reconstruction}
\label{sec:mv_recon}

Given an input video sequence $\{I_t\}$, we reconstruct the scene using Vipe~\cite{huang2025vipe}, which estimates per-frame camera extrinsics $(R_t, \mathbf{t}_t)$ and a dense depth map $D_t$. With camera intrinsics $K$, each pixel $\mathbf{u} = (u, v)$ in frame $t$ with depth $D_t(\mathbf{u})$ is back-projected as
\begin{equation}
    \mathbf{X}_t(\mathbf{u}) = R_t^{-1} \big( D_t(\mathbf{u}) K^{-1} \tilde{\mathbf{u}} - \mathbf{t}_t \big),
\end{equation}
where $\tilde{\mathbf{u}}$ is the homogeneous coordinate of $\mathbf{u}$. Aggregating points from all frames produces a scene-level point cloud $\mathcal{P}$ in a global coordinate system.

\subsection{Point Cloud Bounding-box Normalization}
\label{sec:obb_norm}

We estimate the Oriented Bounding Box (OBB) of $\mathcal{P}$ and obtain its three dimensions with side lengths $l_x, l_y, l_z$. While PCA-based methods~\cite{abdi2010principal} are computationally efficient, they often fail to produce tight bounding boxes as they are driven by data variance rather than geometric bounds. Conversely, exact algorithms based on the rotating calipers paradigm~\cite{o1985finding} guarantee minimal volume but suffer from cubic complexity ($O(n^3)$), rendering them impractical for large-scale point clouds. To address this, we adopt an optimization-based approximation strategy~\cite{chan2001determination}. Specifically, we formulate the orientation estimation as an optimization problem to iteratively refine the bounding box, achieving a balance between computational efficiency and volume minimization.

Upon determining the OBB, we identify the plane defined by the two largest dimensions (assumed to be $l_x$ and $l_y$) as the \emph{candidate} scene plane $XY$. This is based on the empirical observation that in our target scenarios (indoor and driving scenes), the environment layout is generally regular, and the ground (along with the ceiling) typically spans the largest visible area. The remaining axis is assigned as the candidate vertical axis $Z$, noting that its direction (up vs.\ down) remains ambiguous at this stage.

\subsection{Ground-aware Plane Estimation}
\label{sec:ground_plane}

To resolve the $Z$-axis orientation, we exploit a simple ground prior: videos are captured near the ground, so the point cloud contains many ground points but few ceiling points. Consequently, as shown in Fig.~\ref{fig:z_vis}, along the vertical direction the distribution of points tends to exhibit a dense cluster near the ground and a long tail towards the ceiling.

Let $\{z_i\}$ be the coordinates of all points in $\mathcal{P}$ projected onto the candidate $Z$ axis, and let $z^{(0)}, z^{(5)}, z^{(95)}, z^{(100)}$ denote the empirical $0$-th, $5$-th, $95$-th and $100$-th percentiles of $\{z_i\}$, respectively. We then measure the spans of the lowest and highest $5\%$ segments along the candidate $Z$ axis:
\begin{equation}
    d_{\text{bottom}} = z^{(5)} - z^{(0)}, \quad
    d_{\text{top}}    = z^{(100)} - z^{(95)}.
\end{equation}
Due to the long-tail behavior towards the ceiling, the ground side typically forms a much tighter cluster, and we empirically observe $d_{\text{bottom}} \ll d_{\text{top}}$ in most cases. We therefore identify the side with the smaller $5\%$ span (i.e., $\min(d_{\text{bottom}}, d_{\text{top}})$) as the ground side and fix the orientation of the $Z$ axis accordingly. This yields a ground-aligned 3D coordinate system in which the $XY$ plane is parallel to the estimated ground.

\subsection{BEV Representation Generation}
\label{sec:bev}

With the ground-aligned coordinates, we project all points in $\mathcal{P}$ onto the $XY$ plane and rasterize the plane into a regular grid. For each cell, we aggregate the points inside and encode simple geometric statistics (e.g., occupancy or height). This yields a BEV feature map with a consistent metric scale and orientation, which we use as a geometric prior for the downstream network.

\section{Hyper-parameter Settings of SPF-DPP}

SPF-DPP introduces a small set of hyper-parameters controlling geometry, diffusion, and semantic modulation.
We fix all hyper-parameters across all experiments and do not tune them per dataset.
The translation scale is set to $\sigma_t = 1$, and we weight rotational differences more strongly with $\beta = 2$.
The temporal bandwidth is set to $b = 24$, corresponding to roughly one second for videos recorded at 24\,FPS, which yields a sparse yet locally connected viewpoint graph.
For the diffusion kernel, we use a heat scale of $\tau = 2$; since the resulting matrices are moderate in size, we compute $\exp(-\tau \mathcal{L})$ directly without polynomial approximation.
For semantic calibration, we adopt a temperature $T = 1$ and set the mixing weight of the quality to $\alpha = 0.5$, balancing semantic relevance and diversity.
We select a fixed subset size of $k = 32$ frames, following the setting used in prior work for fair comparison.
All hyper-parameters are summarized in Table~\ref{tab:spf_dpp_hparams}. The sensitivity analysis for these hyper-parameters is provided in the main paper (Tab.~\ref{tab:spf_hyper}).

\begin{table}[t]
    \centering
    \footnotesize
    \begin{tabularx}{\linewidth}{>{\centering\arraybackslash}p{2em} >{\raggedright\arraybackslash}X >{\centering\arraybackslash}p{3em}}
        \toprule
        Symbol & Description & Value \\
        \midrule
        $\sigma_t$ & Translation scale (meters normalization) & $1$ \\
        $\beta$    & Rotation weight (rad vs.\ meters)        & $2$ \\
        $b$        & Temporal bandwidth (frames)              & $24$ \\
        $\tau$     & Heat diffusion scale                      & $2$ \\
        $T$        & Softmax temperature for semantic scores   & $1$ \\
        $\alpha$   & Quality mixing weight                     & $0.5$ \\
        $k$        & Selected subset size (frames)             & $32$ \\
        \bottomrule
    \end{tabularx}
    \caption{Hyper-parameters used in SPF-DPP. All values are fixed across experiments.}
    \label{tab:spf_dpp_hparams}
\end{table}

\section{Prompt}

\subsection{System Prompt}
\label{subsec:sp}

\begin{tcolorbox}[
  colback=codegray,
  colframe=black,
  title=\texttt{SYSTEM\_PROMPT},
  fontupper=\ttfamily\small,
  sharp corners,
  boxrule=0.5pt,
  enhanced,
  breakable,
]
\begin{lstlisting}[basicstyle=\small\ttfamily, breaklines=true]
You are a helpful assistant.
# Tools
You may call one or more functions to assist with the user query.
You are provided with function signatures within <tool_call></tool_call> XML tags:
<tool_call>
{"type":"function","function":{"name":"video_image_sample_tool","description":"Given camera parameters in a bird's-eye-view (BEV) coordinate system, this tool generates a realistic and clear video frame to assist in answering the question.","parameters":{"type":"object","properties":{"camera":{"type":"array","items":{"type":"number"},"minItems":3,"maxItems":3,"description":"The camera parameters as [x1, y1, r], where (x1, y1) represents the location on the x-axis and y-axis, and r represents the camera direction: r=270 faces right, r=90 faces left, r=0 faces up, and r=180 faces down in the BEV frame."}}},"required":["camera"]}}}
</tool_call>

# How to call a tool
Return a JSON object containing the function name and arguments within <tool_call></tool_call> XML tags:
<tool_call>
{"name": <function-name>, "arguments": <args-json-object>}
</tool_call>

Example: 
<tool_call> 
{"name": "video_image_sample_tool", "arguments": {"camera": [100, 200, 145]}} 
</tool_call>
\end{lstlisting}
\end{tcolorbox}

\subsection{User Prompt}
\label{subsec:up}

\begin{tcolorbox}[
  colback=codegray,
  colframe=black,
  title=\texttt{USER\_PROMPT},
  fontupper=\ttfamily\small,
  sharp corners,
  boxrule=0.5pt,
  enhanced,
  breakable,
]
\begin{lstlisting}[basicstyle=\small\ttfamily, breaklines=true]
Question: {}

Think first. If needed, call video_image_sample_tool to obtain a real and clear video frame, then answer. 
The output format must strictly be:  <think>...</think>  <tool_call>...</tool_call> (if tools are needed) or <think>...</think> <answer>...</answer>
\end{lstlisting}
\end{tcolorbox}

\end{document}